\documentclass{article}

\newif\ifpreprintversion
\preprintversiontrue       

\newif\ifshowcorrespondence
\showcorrespondencetrue

\usepackage{microtype}
\usepackage{graphicx}
\usepackage{booktabs}
\usepackage{multirow}
\usepackage{hyperref}

\ifpreprintversion
  \usepackage[preprint]{icml2026}
\else
  \usepackage{icml2026}
\fi

\usepackage{amsmath}
\usepackage{amssymb}
\usepackage{mathtools}
\usepackage{amsthm}
\usepackage[capitalize,noabbrev]{cleveref}

\theoremstyle{plain}

\theoremstyle{definition}

\theoremstyle{remark}

\usepackage{colortbl}
\definecolor{tancolor}{RGB}{187,130,90}
\definecolor{slateblue}{RGB}{85,104,154}
\definecolor{lm_purple}{RGB}{227,227,240}
\definecolor{lm_purple_low}{RGB}{240,240,248}

\icmltitlerunning{MoE-nD: Per-Layer Routing for Multi-Axis KV Compression}

\begin{document}

\twocolumn[
  \icmltitle{MoE-nD: Per-Layer Mixture-of-Experts Routing \\
             for Multi-Axis KV Cache Compression}

  \icmlsetsymbol{equal}{*}

  \ifpreprintversion
    \begin{icmlauthorlist}
      \icmlauthor{Libo Sun}{auburn}
      \icmlauthor{Peixiong He}{auburn}
      \icmlauthor{Po-Wei Harn}{ncu}
      \icmlauthor{Xiao Qin}{auburn}
    \end{icmlauthorlist}
    \icmlaffiliation{auburn}{Department of Computer Science and Software Engineering, Auburn University, Auburn, AL, USA}
    \icmlaffiliation{ncu}{Department of Information Management, National Central University, Taoyuan, Taiwan}
    \icmlcorrespondingauthor{Xiao Qin}{xqin@auburn.edu}
  \fi

  \icmlkeywords{KV Cache, Long Context, Compression, Mixture of Experts,
                Per-Layer Routing, Quantization}

  \vskip 0.3in
]

\printAffiliationsAndNotice{}

\begin{abstract}
KV cache memory is the dominant bottleneck for long-context LLM inference.
Existing compression methods each act on a \emph{single} axis of the four-dimensional KV
tensor---token eviction (sequence), quantization (precision), low-rank projection (head
dimension), or cross-layer sharing---but apply the same recipe to every layer.
We show that this homogeneity leaves accuracy on the table: different layers
respond very differently to each compression operation, and the optimal
per-layer mix of eviction and quantization is far from uniform.
We propose \textbf{MoE-nD}, a mixture-of-experts framework that routes each
layer to its own (eviction-ratio, K-bits, V-bits) tuple under a global memory
budget. An offline-calibrated greedy solver chooses the routing that minimizes
predicted quality loss; at inference time, per-layer heterogeneous eviction
and quantization are applied jointly through a single attention patch.
On a 4-task subset of LongBench-v1 (16k inputs, $n=50$ per task,
adapted reasoning-model protocol; see \S\ref{sec:experiments}), MoE-nD's
hetero variant matches our uncompressed 1.9~GB baseline at $14\times$
compression (136~MB) while every other compressed baseline we tested
(1d, 2d\_uniform, 2d) at comparable or smaller memory stays under
$8/100$.
The gains hold on AIME reasoning benchmarks ($+6$ to $+27$pts over the
strongest per-layer-quantization baseline across eight configurations).
Two null results---MATH-500 and LongBench's TREC---share a principled
cause (short inputs, solver picks keep=1.0 on most layers), cleanly
characterizing when per-layer eviction routing has headroom to help.
\end{abstract}

\section{Introduction}
\label{sec:intro}

Long-context reasoning in modern LLMs pushes KV cache memory to tens of GB
per sequence, dominating latency and causing out-of-memory failures on
commodity hardware~\cite{wei2022cot,deepseek2025r1}. KV cache compression
addresses this by reducing either the number of stored tokens
(eviction~\cite{zhang2023h2o,li2024snapkv,cai2025rkv}), the bits per element
(quantization~\cite{liu2024kivi,hooper2024kvquant,yang2024minikv}), or a
mix of the two~\cite{dong2024qaq}. These approaches are almost always applied
uniformly: every transformer layer gets the same budget, the same eviction
ratio, and the same bit-width.

\begin{figure}[t]
  \vskip 0.1in
  \begin{center}
    \includegraphics[width=\columnwidth]{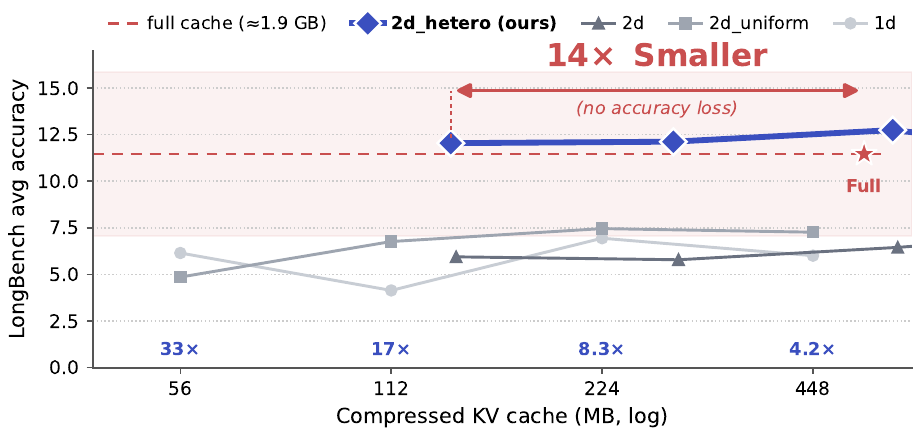}
    \caption{\textbf{LongBench-v1 accuracy vs.\ compressed KV memory on DeepSeek-R1-Distill-Qwen-7B.} 4-task subset (NarrativeQA, HotpotQA, TREC, PassageRetrieval-en), $n=50$ per task, 16k-token inputs (middle truncation), adapted reasoning-model protocol (max\_gen $\times 16$, scoring after final \texttt{</think>}); see \S\ref{sec:experiments}. MoE-nD's hetero variant ($2d_\textrm{hetero}$) matches the uncompressed full-cache baseline at its leftmost operating point (136~MB, $14\times$ compression), while single-axis eviction ($1d$), uniform two-axis ($2d_\textrm{uniform}$), and per-layer-quantization-only ($2d$) lose 33--64\% of LongBench-average accuracy across budgets, and up to 99\% on individual tasks (Table~\ref{tab:longbench-per-task}). The 1d/2d\_uniform points sit at 56--448~MB and the 2d/2d\_hetero points at 136--1175~MB --- the per-method memory differs because the greedy solver spends the 2d quantization headroom on fewer-compressed layers rather than deeper compression (see Table~\ref{tab:longbench-avg}).}
    \label{fig:teaser}
  \end{center}
  \vskip -0.2in
\end{figure}

Recent per-layer work has begun to challenge the uniformity assumption in a
limited way: AdaKV~\cite{feng2025adakv} and PyramidKV~\cite{cai2024pyramidkv}
route \emph{budgets} per layer within eviction, KVTuner~\cite{li2025kvtuner}
routes \emph{quantization bits} per layer. However, to our knowledge, no prior
work routes \emph{across multiple compression axes jointly} at per-layer
granularity. This is a missed opportunity: different layers respond
very differently to eviction vs. quantization, and the optimal allocation
trades the two per layer.

\paragraph{Contribution.}
We propose \textbf{MoE-nD}, a mixture-of-experts~\cite{shazeer2017moe}
formulation that treats each compression axis (eviction ratio,
K-precision, V-precision) as an expert and lets an offline-calibrated
router select the mix per layer under a single global memory budget. The router is a greedy solver over an
offline-measured per-layer sensitivity table; inference-time overhead is
comparable to a uniform two-axis (eviction $+$ quantization) baseline,
not to the cheaper eviction-only single-axis baseline.

Empirically, on a 4-task LongBench-v1 subset (16k inputs, adapted
reasoning-model protocol; see \S\ref{sec:experiments}), MoE-nD's hetero
variant matches the uncompressed 1.9~GB baseline at $14\times$
compression (136~MB) while every comparable method at the same or
smaller memory stays under $8/100$, and it beats the strongest non-hetero two-axis
baseline by $+6$ to $+27$pts across all eight budget$\times$dataset
configurations of AIME-24 and AIME-25. An ablation
chain ($2d_\textrm{uniform} \!\to\! 2d \!\to\! 2d_\textrm{hetero}$)
isolates the contribution of each routing axis and shows that per-layer
\emph{quantization} routing alone adds ${\approx}\,0$pts on average---the
novel lift comes entirely from per-layer \emph{eviction} routing. Two null
results, short-context MATH-500 and LongBench's TREC, share a principled
mechanism (at loose budgets the solver picks keep$=\!1.0$ for most layers,
so hetero degenerates to uniform and no advantage is possible) and let us
cleanly scope the method.

\paragraph{A note on absolute F1.}
\label{par:absolute-scores}
Our primary model is DeepSeek-R1-Distill-Qwen-7B~\cite{deepseek2025r1},
chosen to match the calibration pipeline of our per-layer sensitivity table.
Reasoning models emit long \texttt{<think>\dots</think>} chains before the
answer, so at the LongBench default $\texttt{max\_gen}\in\{32,\ldots,128\}$
tokens, the model almost never reaches an answer. We evaluate with
$\texttt{max\_gen}\times 16$ and score only the text after the final
\texttt{</think>}. Even with these fixes, our uncompressed baseline lands
at 11.5\% LongBench average---well below published Llama-3-8B-Instruct
baselines in the 35--50 range~\cite{li2024snapkv,cai2024pyramidkv}. This gap
comes from reasoning-style generation (verbose, thought-heavy), not from the
harness. All methods in this paper run through the identical pipeline
with identical model, identical prompts, and identical extraction, so the
scientific claim is the \emph{relative} gap between methods at matched
memory. Appendix~\ref{app:absolute-scores} discusses porting to instruction-tuned
models.

\section{Background and Related Work}
\label{sec:related}

The KV cache at layer $\ell$ stores tensors $K_\ell, V_\ell \in
\mathbb{R}^{H_\textrm{kv} \times T \times d_\textrm{head}}$, where $T$ is
sequence length, $H_\textrm{kv}$ is the number of key-value heads
(GQA~\cite{ainslie2023gqa}), and $d_\textrm{head}$ is per-head dimension. Compression methods act on one of
four axes:

\textbf{Eviction (sequence axis).}
StreamingLLM~\cite{xiao2024efficient} retains attention sinks plus a sliding
window. H2O~\cite{zhang2023h2o} and SnapKV~\cite{li2024snapkv} accumulate
attention-based importance scores to identify ``heavy hitters.''
Scissorhands~\cite{liu2023scissorhands} and R-KV~\cite{cai2025rkv} extend
this for long generation. TriAttention~\cite{mao2026triattention} uses
pre-RoPE trigonometric structure for importance estimation. Despite their
differences, these methods all apply a \emph{single} eviction policy to
\emph{every} layer.

\textbf{Per-layer eviction budgets.}
AdaKV~\cite{feng2025adakv} and PyramidKV~\cite{cai2024pyramidkv} vary the
eviction budget per layer (typically deeper layers get smaller budgets).
However they still quantize (if at all) uniformly.

\textbf{Quantization (precision axis).}
KIVI~\cite{liu2024kivi} and KVQuant~\cite{hooper2024kvquant} quantize K and V
to INT4 or INT8 uniformly across layers.  MiniKV~\cite{yang2024minikv}
combines eviction with 2-bit quantization.  KVTuner~\cite{li2025kvtuner}
routes K/V bit-widths per layer, but keeps sequence length fixed
(no eviction).

\textbf{Low-rank and cross-layer.}
ASVD~\cite{yuan2024asvd} projects KV to lower rank. CLA~\cite{brandon2024reducing}
shares KV across adjacent layers. Both are orthogonal to the axes routed
in MoE-nD.

\medskip
\noindent
To our knowledge, no prior method jointly routes \emph{both} eviction ratio
and K/V quantization bits on a per-layer basis under a global memory budget.
MoE-nD fills this gap; the ablation in \S\ref{subsec:ablation} demonstrates
the per-layer eviction routing (not per-layer quantization routing) is the
critical new lever.

\section{Observation: Layers Are Not Equal}
\label{sec:observation}

Before proposing the method, we document the empirical property that
motivates per-layer routing: for the most aggressive compression
operations (high-ratio eviction and K-quantization), the accuracy cost
varies by two to nearly three orders of magnitude across transformer
layers of DeepSeek-R1-Distill-Qwen-7B; for V-quantization and mild
eviction, the variation is smaller but the trade-off between axes still
flips per layer.

\begin{figure*}[t]
  \vskip 0.05in
  \begin{center}
    \includegraphics[width=0.96\textwidth]{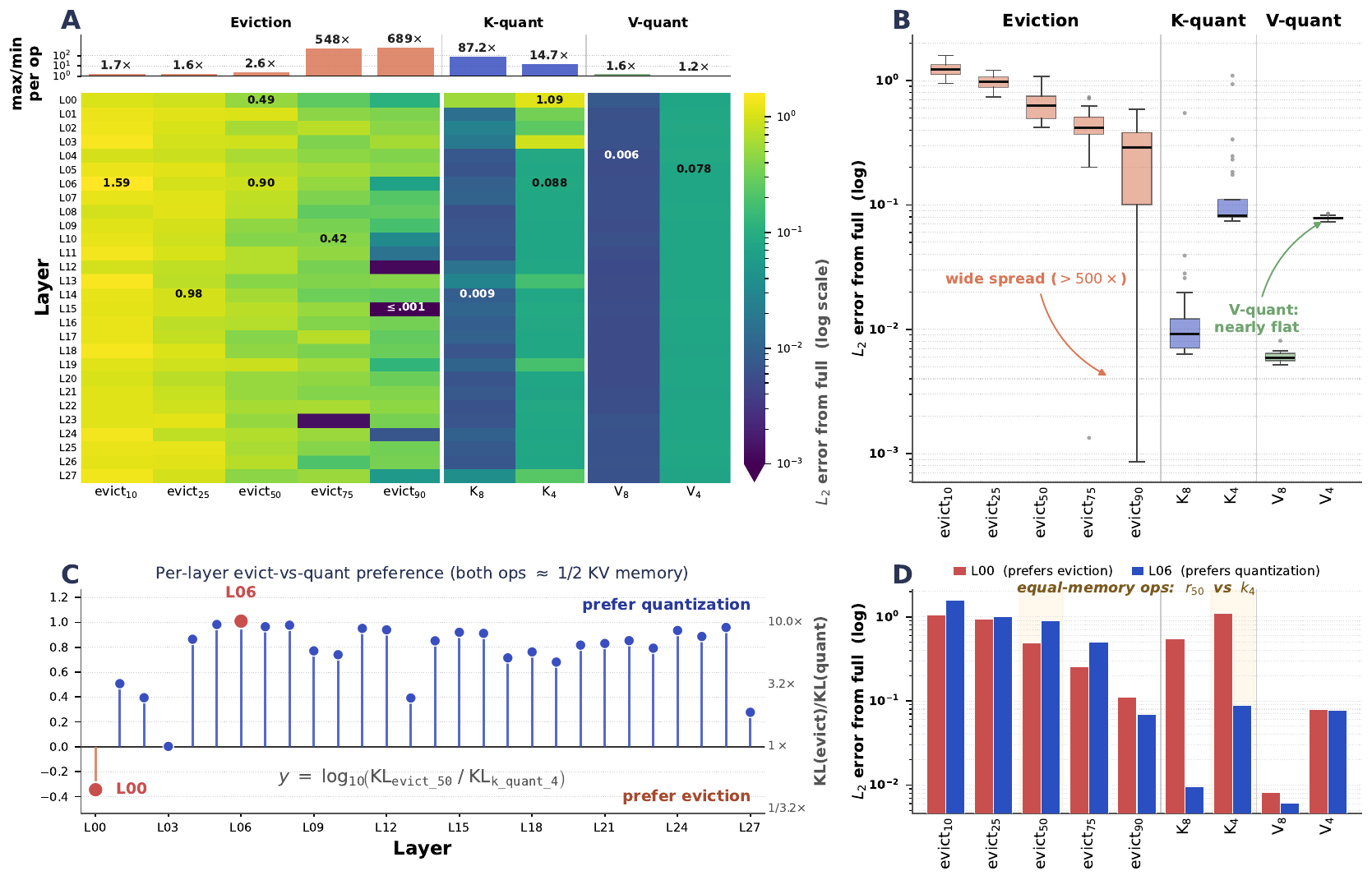}
    \caption{\textbf{Per-layer sensitivity landscape.}
    \emph{(A)} Full $28\times 9$ heatmap of predicted attention-output
    $L_2$ error when applying each compression operation to a single layer,
    measured offline against the uncompressed reference (log-scale color;
    only headline cells are annotated to keep the panel readable). The top
    strip shows per-operation $\max_\ell\!/\!\min_\ell$ ratio, directly
    summarizing the heterogeneity numbers in
    Table~\ref{tab:sensitivity-stats}: aggressive eviction (evict$_{75}$,
    evict$_{90}$) spans $500\text{--}700\times$ across layers while V-quant
    spans only $1.2\text{--}1.6\times$.
    \emph{(B)} Per-op error distribution across layers as box plots:
    V-quant boxes are nearly flat (uniform-OK) while evict$_{75}/90$
    sprawl across three orders of magnitude.
    \emph{(C)} Per-layer evict-vs-quant preference at equal memory cost
    (evict$_{50}$ vs K-quant-4, both $\approx$1/2 of full precision).
    Positive bars: eviction damages more than quant on this layer
    (prefer quant); negative bars: quant damages more (prefer eviction).
    L00 is the lone layer preferring eviction; the routing solver must
    discover this without a uniform heuristic.
    \emph{(D)} L00 vs L06 head-to-head: $L_2$ error across all 9 ops,
    with the equal-memory pair (evict$_{50}$ vs $k_4$) shaded --- concrete
    proof of the ``two layers, opposite preferences'' phenomenon cited
    in the text.}
    \label{fig:sensitivity}
  \end{center}
  \vskip -0.15in
\end{figure*}

We measure, for each layer $\ell$ and each candidate compression
configuration $c = (\textrm{keep\_ratio}, k_\textrm{bits}, v_\textrm{bits})$,
the relative $L_2$ error between the layer's full-precision attention
output and its compressed-attention output (with $c$ applied to layer
$\ell$ only),
$S_{\ell,c} = \|\mathrm{attn}_{\mathrm{full}}^{(\ell)} - \mathrm{attn}_c^{(\ell)}\|_2 \,/\, \|\mathrm{attn}_{\mathrm{full}}^{(\ell)}\|_2$,
evaluated on a single 27-token reasoning prompt. We use this lightweight
proxy as our sensitivity score; we validated it post hoc against a
principled KL-divergence calibration over 8 held-out sequences of 2048
tokens (Appendix~\ref{app:implementation}) and find a mean per-layer
Pearson correlation of $0.945$ (Spearman $0.937$) across all 28 layers
and 11 configs --- i.e.\ the proxy produces the same within-layer
rankings the greedy solver consumes. Three empirical properties of the
resulting sensitivity table $S \in \mathbb{R}^{L \times |\mathcal{C}|}$
(Figure~\ref{fig:sensitivity}, with numerical summaries in
Table~\ref{tab:sensitivity-stats}) jointly motivate per-layer joint
routing.

\begin{table}[t]
\caption{Per-operation variation across 28 layers of DeepSeek-R1-Distill-Qwen-7B, measured as $\max_\ell s_\ell / \min_\ell s_\ell$ where $s_\ell$ is the relative attention-output $L_2$ error from applying the operation to layer $\ell$ alone. Large ratios indicate strong layer heterogeneity --- exactly the regime where per-layer routing can help.}
\centering
\small
\begin{tabular}{l|rr|r}
\toprule
\textbf{Operation} & \textbf{min $s_\ell$} & \textbf{max $s_\ell$} & \textbf{max/min} \\
\midrule
evict\_10\%   & 0.95   & 1.59   & 1.7$\times$ \\
evict\_25\%   & 0.73   & 1.21   & 1.6$\times$ \\
evict\_50\%   & 0.42   & 1.08   & 2.6$\times$ \\
evict\_75\%   & 0.0013 & 0.73   & \textbf{548}$\times$ \\
evict\_90\%   & 0.0009 & 0.59   & \textbf{689}$\times$ \\
k\_quant\_8   & 0.0063 & 0.55   & 87$\times$ \\
k\_quant\_4   & 0.075  & 1.09   & 15$\times$ \\
v\_quant\_8   & 0.0052 & 0.0081 & 1.6$\times$ \\
v\_quant\_4   & 0.073  & 0.085  & 1.2$\times$ \\
\bottomrule
\end{tabular}
\label{tab:sensitivity-stats}
\end{table}

\paragraph{Eviction sensitivity diverges sharply at aggressive ratios.}
At mild eviction ($\leq 50\%$), the most and least sensitive layers differ
by only $1.6$--$2.6\times$, so a uniform policy is roughly defensible. At
aggressive eviction ($75\%$, $90\%$) the ratio explodes to $548\times$
and $689\times$ respectively, and per-layer routing becomes indispensable.

\paragraph{K-quantization varies strongly; V-quantization does not.}
K-quant at 4 bits spans a $15\times$ range across layers (at 8 bits,
$87\times$), while V-quant at either bit-width spans only
$1.2\text{--}1.6\times$. Per-layer K-quant routing therefore has real
headroom; V-quant routing is essentially free to apply uniformly.

\paragraph{Eviction and quantization trade at different rates per layer.}
Comparing equal-memory operations (\texttt{evict\_50\%} vs
\texttt{k\_quant\_4}), on L00 eviction is $2.2\times$ cheaper than quant
($0.49$ vs $1.09$), whereas on L06 quant is $10\times$ cheaper
($0.088$ vs $0.90$). The trade-off is not globally consistent, so the
choice of \emph{which} axis to compress on a given layer is itself a
per-layer decision.

The first two properties show that each compression axis has
per-layer structure --- a \emph{single-axis} per-layer router (AdaKV for
eviction, KVTuner for quant) can exploit one axis but not both. The third
property shows the two axes are \emph{entangled}: choosing which to apply on
each layer requires a joint router that sees both. This is the empirical
foundation of MoE-nD.

\section{MoE-nD}
\label{sec:method}

\begin{figure*}[t]
  \vskip 0.2in
  \begin{center}
    \includegraphics[width=0.95\textwidth]{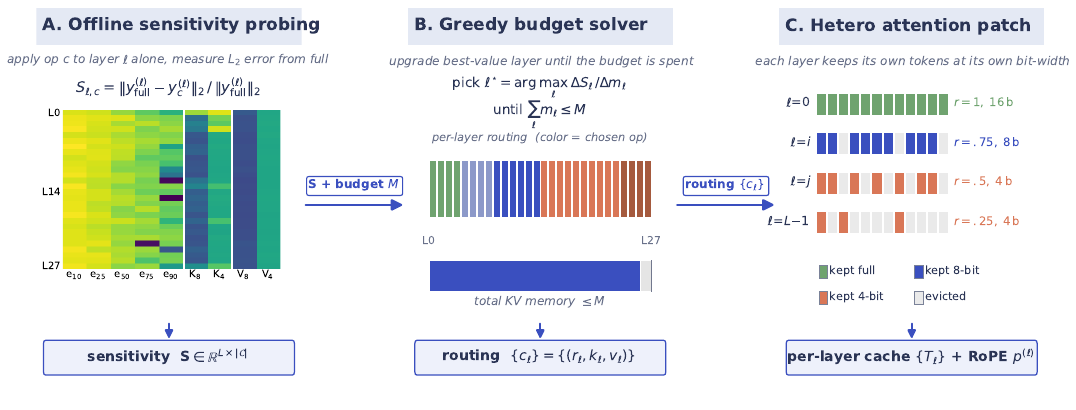}
    \caption{\textbf{MoE-nD pipeline.}
    \emph{(A)} Offline, per-layer sensitivity to each compression operation
    is measured against an uncompressed reference; the heatmap shows the
    actual calibration table for DeepSeek-R1-Distill-Qwen-7B.
    \emph{(B)} A greedy budget solver picks the per-layer
    $(\textrm{keep}, k_\textrm{bits}, v_\textrm{bits})$ tuple that
    greedily reduces predicted total sensitivity (attention-output $L_2$
    error) subject to a global memory budget $M$.
    \emph{(C)} At inference time, an attention patch applies the routed
    eviction and quantization to each layer jointly, with per-layer RoPE
    re-inversion. Routing colors in (B) and keep-patterns in (C) are
    illustrative; the heatmap in (A) is real data.}
    \label{fig:method-overview}
  \end{center}
  \vskip -0.2in
\end{figure*}

Figure~\ref{fig:method-overview} summarizes the pipeline: an offline
calibration phase measures $S$ once per model; a greedy budget solver
produces a per-layer routing; at inference time, the routed eviction and
quantization are applied jointly through a single attention patch.

\paragraph{Design space.}
Each layer exposes three orthogonal knobs: a keep ratio
$\textrm{keep\_ratio}_\ell \in \{0.1, 0.25, 0.5, 0.75, 0.9, 1.0\}$ that
sets the fraction of cache tokens retained after eviction (scored by a
TriAttention-style trigonometric importance
signal~\cite{mao2026triattention}), a K bit-width
$k_{\textrm{bits},\ell} \in \{16, 8, 4\}$ applied per channel, and a V
bit-width $v_{\textrm{bits},\ell} \in \{16, 8, 4\}$ applied with
asymmetric group quantization. We intentionally exclude low-rank projection
and cross-layer KV sharing, which we found in internal experiments do not
compose cleanly with GQA-style sensitivity routing on dense models.

\paragraph{Memory cost.}
At layer $\ell$ the compressed KV size is
\begin{equation}
\begin{aligned}
  m_\ell(c_\ell) = \ & \underbrace{\textrm{keep\_ratio}_\ell \cdot T_\textrm{cache}}_{\textrm{tokens}}
                     \cdot H_\textrm{kv} \cdot d_\textrm{head} \\
                   & \cdot \frac{k_{\textrm{bits},\ell} + v_{\textrm{bits},\ell}}{8}
                     \ \ \text{bytes},
\end{aligned}
\end{equation}
and the solver must satisfy $\sum_\ell m_\ell(c_\ell) \leq M$ for a
user-specified global budget $M$.

\paragraph{Greedy solver.}
\label{subsec:solver}
Finding the globally optimal $\{c_\ell\}$ is a constrained discrete problem
with $(6 \cdot 3 \cdot 3)^{L}$ candidates --- on a 28-layer model,
${\approx}\,10^{42}$ --- and exact search is intractable. We instead use
the same greedy allocator as KVTuner~\cite{li2025kvtuner} but apply it to
the full three-axis table: starting from the cheapest configuration on
every layer, we iteratively upgrade the layer whose marginal sensitivity
reduction per unit memory ($\Delta S_\ell / \Delta m_\ell$) is largest,
until $M$ is exhausted. Solve time is $<50$ms on CPU.

\paragraph{Heterogeneous attention patch.}
The per-layer heterogeneous cache creates an implementation challenge:
every layer has a different cache length
$T_\ell = \textrm{keep\_ratio}_\ell \cdot T_\textrm{cache}$, and positions
diverge after the first eviction round. We address this with a small
modification of the attention kernel: per-layer cache position arrays
$p^{(\ell)} \in \mathbb{Z}^{T_\ell}$ track the original positions of the
tokens each layer retained, RoPE~\cite{su2024roformer} inversion uses
$p^{(\ell)}$ (not the global position) when re-rotating during the next
generation step, and
eviction fires on a step-count trigger (every $\beta = 128$ generated
tokens) to avoid runaway costs when some layers have
$\textrm{keep\_ratio}=1.0$ and never evict. This reduces to the standard
attention patch when all layers have identical $T_\ell$, so
uniform-eviction methods are a strict special case.

\section{Experiments}
\label{sec:experiments}

\subsection{Setup}

\textbf{Model.} DeepSeek-R1-Distill-Qwen-7B~\cite{deepseek2025r1} (28 layers,
8 KV heads, head dim 128). All methods evaluated in bfloat16 on a single
NVIDIA H200 with eager attention (compression hooks are incompatible with
FlashAttention~\cite{dao2022flashattention} in our current implementation).

\textbf{Benchmarks.}
(1) \emph{LongBench-v1}~\cite{bai2024longbench} --- 4 tasks chosen to
stress the KV cache (all four exceed our $T_\textrm{cache}=16$k input
window before middle-truncation) while spanning four task formats:
extractive QA (NarrativeQA), multi-hop QA (HotpotQA), classification
(TREC), and retrieval (PassageRetrieval-en). $n=50$ per task; inputs
truncated to 16k tokens (middle truncation). The remaining LongBench
tasks were excluded because they fit comfortably under any reasonable
KV budget on this 7B model and so cannot exercise per-layer routing
(see also \S\ref{subsec:null-results} for the same condition on
short-prompt benchmarks).
(2) \emph{AIME-24} and \emph{AIME-25} --- 30 problems each, 16k-token max
generation.
(3) \emph{MATH-500}~\cite{hendrycks2021math} --- 500 problems; short prompts
(avg $<1\textrm{k}$).

\textbf{Methods.} We compare four MoE-nD variants against a single-axis
baseline. \textbf{full} applies no compression and serves as the
accuracy/memory upper bound. \textbf{1d} is uniform eviction at budget $b$
(TriAttention-style). \textbf{2d\_uniform} adds fixed K8/V4 quantization
on every layer (MiniKV-like), \textbf{2d} replaces the uniform
quantization with per-layer routed K/V bits (KVTuner + TriAttention), and
\textbf{2d\_hetero}---the full MoE-nD proposal---further routes eviction
ratios per layer. Budgets are $\{64, 128, 256, 512\}$ for MATH/AIME (short generations) and
$\{512, 1024, 2048, 4096\}$ for LongBench (long inputs). The 2d/2d\_hetero
methods receive a scaled token budget $b_\textrm{2d} = b \cdot 4/1.5$
intended to give them the \emph{same} memory as 1d at 16-bit when their
quantization knob is fully exercised. In practice the greedy solver
prefers to spend that headroom on extra full-precision layers rather
than uniformly aggressive 4-bit quantization, so per-cell actual memory
ends up higher for 2d/2d\_hetero (e.g.\ at $b=512$: 1d/2d\_uniform use
56~MB, 2d uses 139~MB, 2d\_hetero uses 136~MB). We therefore report
per-method memory alongside accuracy in
Table~\ref{tab:longbench-avg} and compare both at \emph{matched nominal
budget} (where 2d\_hetero competes against 2d at near-identical memory)
and at \emph{matched memory bucket} (where 2d\_hetero at 136~MB is the
only method achieving baseline accuracy).

\textbf{Reasoning-model handling.} As noted in \S\ref{sec:intro},
$\texttt{max\_gen}$ is scaled by $16\times$ vs.\ LongBench defaults and the
text after the final \texttt{</think>} is used for scoring; first-line
truncation is applied to TREC per the LongBench official protocol.

\subsection{Main result: LongBench accuracy vs.\ memory}

Figure~\ref{fig:teaser} and Table~\ref{tab:longbench-avg} show the headline
comparison: average task accuracy as a function of compressed KV memory.

\begin{table}[t]
\caption{LongBench average (over 4 tasks, $n=50$). Methods share a nominal eviction budget $b$ (token count); actual KV memory differs per method family because the solver spends the 2d $4/1.5\times$ headroom on less-compressed layers (\emph{italic MB} sub-rows). At $b=512$, $2d_\textrm{hetero}$ uses 136~MB ($14\times$ compression vs the 1859~MB full cache) and matches the baseline (12.0 vs 11.5); at the comparable 139~MB, $2d$ only reaches 5.9. Compression ratios for the four hetero cells are $14, 6.6, 3.2, 1.6\times$.}
\centering
\scriptsize
\setlength{\tabcolsep}{3pt}
\begin{tabular}{l|cccc|c}
\toprule
                  & b=512 & b=1024 & b=2048 & b=4096 & full \\
\midrule
1d                & 6.1  & 4.1  & 6.9  & 6.0  & --- \\
\textit{\,MB}     & \textit{56}  & \textit{112} & \textit{224} & \textit{448}  & --- \\
2d\_uniform       & 4.9  & 6.8  & 7.5  & 7.7  & --- \\
\textit{\,MB}     & \textit{56}  & \textit{112} & \textit{224} & \textit{448}  & --- \\
2d                & 5.9  & 5.8  & 6.5  & 7.3  & --- \\
\textit{\,MB}     & \textit{139} & \textit{283} & \textit{582} & \textit{1175} & --- \\
\rowcolor{lm_purple}
2d\_hetero (ours) & \textbf{12.0} & \textbf{12.1} & \textbf{12.7} & \textbf{11.5} & --- \\
\rowcolor{lm_purple}
\textit{\,MB}     & \textit{136} & \textit{283} & \textit{582} & \textit{1175} & --- \\
\midrule
full              & ---  & ---  & ---  & ---  & 11.5 \\
\textit{\,MB}     & ---  & ---  & ---  & ---  & \textit{1859} \\
\bottomrule
\end{tabular}
\label{tab:longbench-avg}
\end{table}

$2d_\textrm{hetero}$ at its leftmost operating point (136~MB,
$14\times$ compression) reaches 12.0 vs the uncompressed baseline's 11.5
on the 4-task average ($n=50$ per task; per-task 95\% Wilson CIs are
$\pm 6$--$9$pts). The 0.5pt gap is well inside any reasonable
measurement error at this sample size, so the headline observation is
that $14\times$ compression comes with no detectable accuracy loss in
this protocol --- not that the two are formally equal. The directly
comparable head-to-head is $2d$ at 139~MB (essentially the same memory
as 2d\_hetero at 136~MB), which reaches only 5.9 --- roughly half the
baseline and well outside the CI; per-layer eviction routing is what
buys the remaining 6pts. Single-axis methods at half that memory
(56~MB, $33\times$ compression) reach 4.9--6.1 --- they fail at deeper
compression \emph{and} use less memory, so they are not a clean
apples-to-apples comparison to 2d\_hetero. Breaking it down per task
(Table~\ref{tab:longbench-per-task}), three of the four LongBench tasks
show consistent hetero wins; TREC shows no hetero advantage (discussed
in \S\ref{subsec:null-results}).

\begin{table}[t]
\caption{LongBench per task ($n=50$). Tasks: HQA (HotpotQA), NQA (NarrativeQA), PRE (PassageRetrieval-en), TREC. Metrics: QA-F1 (HQA, NQA), retrieval-score (PRE), exact-match (TREC). Columns are nominal eviction budget $b$; per-method actual memory is in Table~\ref{tab:longbench-avg} and is the same across tasks. \emph{full} is budget-invariant.}
\centering
\scriptsize
\setlength{\tabcolsep}{4pt}
\begin{tabular}{ll|cccc}
\toprule
\textbf{Task} & \textbf{Method} & b=512 & b=1024 & b=2048 & b=4096 \\
\midrule
\multirow{5}{*}{HQA}
  & full        & \multicolumn{4}{c}{9.7} \\
  & 1d          & 0.2 & 0.2 & 0.2 & 1.0 \\
  & 2d\_uniform & 1.1 & 0.1 & 1.4 & 2.4 \\
  & 2d          & 1.1 & 0.3 & 1.4 & 2.4 \\
  \rowcolor{lm_purple}
  & 2d\_hetero  & \textbf{10.9} & \textbf{9.4} & \textbf{10.0} & \textbf{9.7} \\
\midrule
\multirow{5}{*}{NQA}
  & full        & \multicolumn{4}{c}{5.4} \\
  & 1d          & 0.3 & 0.3 & 0.6 & 0.3 \\
  & 2d\_uniform & 0.8 & 0.2 & 0.5 & 1.6 \\
  & 2d          & 0.2 & 0.3 & 0.5 & 1.7 \\
  \rowcolor{lm_purple}
  & 2d\_hetero  & \textbf{7.0} & \textbf{6.4} & \textbf{6.6} & \textbf{5.4} \\
\midrule
\multirow{5}{*}{PRE}
  & full        & \multicolumn{4}{c}{8.7} \\
  & 1d          & 0.1 & 0.1 & 1.0 & 2.7 \\
  & 2d\_uniform & 1.5 & 0.7 & 4.0 & 4.9 \\
  & 2d          & 2.5 & 0.5 & 2.0 & 3.0 \\
  \rowcolor{lm_purple}
  & 2d\_hetero  & \textbf{10.3} & \textbf{10.7} & \textbf{10.3} & 8.7 \\
\midrule
\multirow{5}{*}{TREC}
  & full        & \multicolumn{4}{c}{22.0} \\
  & 1d          & \textbf{24.0} & 16.0 & \textbf{26.0} & 20.0 \\
  & 2d\_uniform & 16.0 & \textbf{26.0} & 24.0 & 22.0 \\
  & 2d          & 20.0 & 22.0 & 22.0 & 22.0 \\
  & 2d\_hetero  & 20.0 & 22.0 & 24.0 & 22.0 \\
\bottomrule
\end{tabular}
\label{tab:longbench-per-task}
\end{table}

\subsection{AIME reasoning benchmarks}

On AIME-24 and AIME-25, MoE-nD $2d_\textrm{hetero}$ beats the non-hetero
two-axis baseline $2d$ at every single budget and dataset
(Table~\ref{tab:aime}). The advantage grows at tighter budgets: at $b=64$
on AIME-25, $2d_\textrm{hetero}$ delivers $30\%$ while every other method
lands at or near $0$.

\begin{table}[t]
\caption{AIME-24 and AIME-25 accuracy ($n=30$). $2d_\textrm{hetero} > 2d$ at every cell; the gap widens at tight budgets where single-axis and uniformly-routed methods collapse to $0$. Individual 95\% CIs at $n=30$ are wide ($\pm 13$--$17$pts), so any single cell is not statistically conclusive; the qualitative pattern (hetero $>$ 2d in 8/8 cells, with the gap widening as budget tightens) is what we rely on. A formal hypothesis test would require independent draws, which these cells (same model and harness) are not.}
\centering
\small
\begin{tabular}{l|cccc}
\toprule
\textbf{Method} & b=64 & b=128 & b=256 & b=512 \\
\midrule
\multicolumn{5}{c}{\textit{AIME-24, full = 26.7}} \\
\midrule
1d         & 0.0 & 0.0 & 0.0 & 10.0 \\
2d\_uniform& 0.0 & 10.0 & 20.0 & 30.0 \\
2d         & 0.0 & 10.0 & 13.3 & 23.3 \\
\rowcolor{lm_purple}
2d\_hetero & \textbf{16.7} & \textbf{20.0} & \textbf{26.7} & \textbf{30.0} \\
\midrule
\multicolumn{5}{c}{\textit{AIME-25, full = 30.0}} \\
\midrule
1d         & 0.0 & 0.0 & 6.7 & 13.3 \\
2d\_uniform& 3.3 & 10.0 & 13.3 & 30.0 \\
2d         & 3.3 & 3.3 & 23.3 & 23.3 \\
\rowcolor{lm_purple}
2d\_hetero & \textbf{30.0} & \textbf{26.7} & \textbf{36.7} & \textbf{33.3} \\
\bottomrule
\end{tabular}
\label{tab:aime}
\end{table}

On AIME-25, $2d_\textrm{hetero}$ at $b=256$ exceeds the full-cache baseline
(36.7 vs 30.0). A similar inversion appears on MATH-500 where $1d$ at
$b=256$ beats full (53.6 vs 50.4). Both happen on reasoning tasks with long
chain-of-thought; the likely mechanism is that aggressive eviction removes
self-distracting intermediate tokens, letting the model re-focus on the
original prompt. Notably, this inversion is visible under both single-axis
and hetero-routed eviction, suggesting it is a property of \emph{eviction
itself} and is orthogonal to MoE-nD's per-layer routing contribution. We
flag it as a phenomenon worth studying but do not claim credit for it in
our main result.

\subsection{Ablation: which routing matters?}
\label{subsec:ablation}

The ablation chain $2d_\textrm{uniform} \to 2d \to 2d_\textrm{hetero}$
cleanly separates the contribution of per-layer quant routing ($\Delta_\textrm{quant}$)
from per-layer eviction routing ($\Delta_\textrm{evict}$), since each
subsequent method adds exactly one routing dimension to the previous.

\begin{table}[t]
\caption{Ablation on AIME-24 and AIME-25. $\Delta_\textrm{quant} = 2d - 2d_\textrm{uniform}$ isolates per-layer quant routing; $\Delta_\textrm{evict} = 2d_\textrm{hetero} - 2d$ isolates per-layer eviction routing. The novel lift comes from eviction routing.}
\centering
\small
\begin{tabular}{ll|rr}
\toprule
\textbf{Dataset} & \textbf{Budget} & $\Delta_\textrm{quant}$ & $\Delta_\textrm{evict}$ \\
\midrule
AIME-24 & 64   & $+0.0$ & $+16.7$ \\
AIME-24 & 128  & $+0.0$ & $+10.0$ \\
AIME-24 & 256  & $-6.7$ & $+13.3$ \\
AIME-24 & 512  & $-6.7$ & $+6.7$ \\
AIME-25 & 64   & $+0.0$ & $+26.7$ \\
AIME-25 & 128  & $-6.7$ & $+23.3$ \\
AIME-25 & 256  & $+10.0$ & $+13.3$ \\
AIME-25 & 512  & $-6.7$ & $+10.0$ \\
\midrule
\textbf{Mean}    &     & $\mathbf{-2.1}$ & $\mathbf{+15.0}$ \\
\bottomrule
\end{tabular}
\label{tab:ablation-chain}
\end{table}

$\Delta_\textrm{evict}$ is positive in 8/8 configurations, averaging
+15.0pts; $\Delta_\textrm{quant}$ is approximately 0 on average
(-2.1pts) and negative in 4 of 8 cells. The same signature holds on
LongBench (Table~\ref{tab:longbench-avg}): across the four budgets,
mean $\Delta_\textrm{evict} = +5.7$pts (positive in 4/4) while
$\Delta_\textrm{quant} = -0.35$pts. \emph{Per-layer eviction routing,
not per-layer quantization routing, is the load-bearing novel
contribution on both reasoning and long-context tasks.}

\subsection{Scope: when does it help?}
\label{subsec:null-results}

Two datasets in our suite show no MoE-nD advantage: MATH-500
(Table~\ref{tab:math500-null}) and LongBench TREC
(Table~\ref{tab:longbench-per-task}, last block). The shared cause is
short prompts under loose budgets. MATH-500 prompts average
${\sim}900$ tokens and TREC prompts average ${\sim}5\textrm{k}$, so in
both cases the uncompressed cache is small relative to the memory
budget. Under these conditions the greedy solver picks
$\textrm{keep\_ratio}=1.0$ for $>\!75\%$ of layers; with few layers
actually evicted, the hetero-eviction routing has nothing to diversify
over and $2d_\textrm{hetero}$ degenerates to $2d$.

\begin{table}[t]
\caption{MATH-500 ($n=500$). $2d_\textrm{hetero}$ does not beat $2d$ on this short-context benchmark because the greedy solver picks keep=1.0 on most layers and the hetero-eviction routing has nothing to diversify over (\S\ref{subsec:null-results}). $1d$ underperforms at tight $b$ where the others get to keep more layers via the quantization headroom; from $b=128$ onward the four compressed methods are within 8pts of each other and of full. We report this negative result explicitly.}
\centering
\small
\begin{tabular}{l|cccc}
\toprule
\textbf{Method} & b=64 & b=128 & b=256 & b=512 \\
\midrule
full       & \multicolumn{4}{c}{50.4} \\
1d         & 16.6 & 40.6 & 53.6 & 57.0 \\
2d\_uniform& 47.4 & 49.4 & 56.0 & 51.4 \\
2d         & 49.0 & 55.6 & 54.4 & 50.6 \\
2d\_hetero & 48.6 & 50.4 & 48.0 & 50.0 \\
\bottomrule
\end{tabular}
\label{tab:math500-null}
\end{table}

This is not a failure of the calibration or the solver---both behave
correctly given the inputs. It is a principled scope condition:
\emph{MoE-nD's hetero variant delivers measurable gains primarily when
the combination of input length and budget tightness forces the solver
to route non-trivially across layers.} On long-context tasks (LongBench, 16k inputs) or long-generation tasks
(AIME at tight $b$), this condition holds everywhere we tested. On
short-context tasks it does not.

A useful corollary for practitioners: in this short-context regime the
simpler $2d_\textrm{uniform}$ implementation suffices, because the
$2d_\textrm{hetero}$ solver would have routed most layers to
nearly-uniform allocations anyway.

\section{Limitations and Future Work}
\label{sec:limitations}

\textbf{External attention-eviction baselines.} Direct comparisons to
SnapKV~\cite{li2024snapkv} and H2O~\cite{zhang2023h2o} at matched memory
are deferred to a follow-up. Both are single-axis attention-based eviction
methods and we expect them to perform in the band of our $1d$ baseline,
which uses the same eviction mechanism with a trigonometric importance
signal in place of attention scoring. Since $1d$ already underperforms
$2d_\textrm{hetero}$ by $\geq 5$pts at every LongBench budget, the
methodological contribution of joint per-layer routing is unlikely to be
overturned by adding these single-axis comparators.

\textbf{Lightweight calibration proxy (validated).} Our sensitivity
table is built from a single 27-token reasoning prompt scored by
attention-output $L_2$ error. We validated this proxy against a
principled KL-divergence calibration over 8 held-out sequences of 2048
tokens (Appendix~\ref{app:implementation}); per-layer Pearson
correlation is $0.945$ on average and $> 0.8$ in 28/28 layers. The
solver consumes within-layer rankings, which are preserved. The
remaining caveat is that \emph{cross-layer} rankings under random
eviction are noisy across calibration metrics (Pearson $r \approx 0$);
a deterministic, attention-aware eviction signal would tighten this
and is a natural next step.

\textbf{Single model family.} Our sensitivity table is calibrated for
DeepSeek-R1-Distill-Qwen-7B. Porting to Llama-3 / Mistral families
requires re-calibration ($\sim$1 GPU-day per family). We expect the
per-layer heterogeneity observed in \S\ref{sec:observation} to hold
broadly---the mechanism (RoPE and GQA structure, MLP residual scales) is
architecture-agnostic---but this is an empirical question.

\textbf{Eager attention.} The heterogeneous attention patch is not yet
compatible with FlashAttention. Peak memory during prefill is
$O(T^2)$ for the attention matrix, which dominates the compressed KV size
at $T=16$k. This limits the practical deployment memory advantage
until a FlashAttention-compatible implementation lands.

\textbf{Instruction-tuned baselines.} Absolute F1 numbers in this paper
reflect reasoning-style generation. A port to a non-reasoning variant
(e.g., Qwen2.5-7B-Instruct) would make our numbers directly comparable to
published LongBench tables; the relative gaps between methods should be
preserved.

\section{Conclusion}
\label{sec:conclusion}

We presented MoE-nD, a KV cache compression method that jointly routes
eviction ratio and K/V quantization bits per layer under a global memory
budget. On long-context benchmarks where per-layer heterogeneity exists,
MoE-nD's full hetero variant matches the uncompressed baseline at
$14\times$ compression (136~MB vs the 1.9~GB full cache) and, at
matched memory, dominates the uniform two-axis baseline by
$1.6$--$2\times$ in F1 across the four budgets we tested. An ablation
chain establishes that per-layer eviction routing, not per-layer
quantization routing, is the novel lever. Two null results---MATH-500 and
LongBench TREC---share a principled cause (short inputs, solver picks
keep=1.0) and cleanly scope the method. We believe MoE-nD provides a
template for compression methods that combine multiple orthogonal axes
under a single calibrated router.

\bibliographystyle{icml2026}
\bibliography{main}

\appendix
\onecolumn

\section{Absolute Scores and Instruction-Tuned Models}
\label{app:absolute-scores}

Our 11.5 LongBench average for the uncompressed baseline is
${\approx}\,3\times$ below Llama-3-8B-Instruct baselines published in
SnapKV~\cite{li2024snapkv} and PyramidKV~\cite{cai2024pyramidkv}. The gap
originates in three compounding factors. First, DeepSeek-R1-Distill is a
reasoning-distilled model: for NarrativeQA it emits extensive reasoning
inside \texttt{<think>\dots</think>} before the final answer, and when
\texttt{max\_gen} is exhausted mid-think, no answer is emitted at all.
Second, even with $\texttt{max\_gen}\times 16$ the final answer is often
verbose and scores poorly against LongBench's concise-gold-answer F1
protocol. Third, we use a 7B reasoning distillation; the published
baselines use 8B instruction-tuned models with 10--20\% more parameters
and task-aligned finetuning. We have not ported to an instruction-tuned
variant primarily because our per-layer sensitivity table is calibrated
for the reasoning model; we plan a Qwen2.5-7B-Instruct port in a revision.

\section{Implementation Details}
\label{app:implementation}

\textbf{Sensitivity calibration.} A single 27-token mathematical-reasoning
prompt (``Solve the equation $x^2 - 5x + 6 = 0$. Think step by step and
show all work.''). For each layer $\ell$ and configuration $c \in C$, we
apply $c$ to layer $\ell$ only and measure
$S_{\ell,c} = \|\mathrm{attn}_{\mathrm{full}}^{(\ell)} - \mathrm{attn}_c^{(\ell)}\|_2 \,/\, \|\mathrm{attn}_{\mathrm{full}}^{(\ell)}\|_2$,
the relative $L_2$ error of the layer's attention output against the
full-precision reference, averaged over the prompt's token positions.
This is a deliberately lightweight proxy: total calibration cost is
seconds-to-minutes on a single H200, the persisted table
\texttt{calibration/ds\_qwen7b\_sensitivity.pt} is 5~kB, and the same
table drives every routing decision in the paper.

\textbf{KL-divergence validation of the proxy.} We validated the
$L_2$-error proxy against a principled KL-divergence calibration
over 8 held-out sequences of 2048 tokens
(\texttt{calibration/ds\_qwen7b\_sensitivity\_kl.pt}, 5~kB). Across the
28 layers and 11 compression configurations, the mean per-layer Pearson
correlation between the two metrics is $0.945$ (Spearman $0.937$); 28/28
layers show Pearson $r > 0.8$, most $> 0.9$. This confirms that the
cheap proxy produces \emph{equivalent within-layer rankings} to the
gold-standard KL method --- and within-layer rankings are exactly what
the greedy budget solver consumes
(\S\ref{subsec:solver}: at each step, ``upgrade layer $\ell$ to its
next-best config''). Per-config cross-layer rankings are tighter for
quantization configs (Pearson $r = 0.99$ for $k_8 v_4$, $0.98$ for
$k_8$, $0.73$ for $k_4$) than for V-quantization (small effect anyway:
$r = 0.01$--$0.41$) or eviction configs ($r \approx 0$); the eviction
disagreement is a property of \emph{random-permutation eviction}
itself --- different random seeds rank the layers differently --- and
does not affect solver decisions, which operate within-layer.

\textbf{Greedy solver.} Python, $<50$ms per invocation. Produces a
\texttt{LayerCompressionConfig} per layer stored on the patched model as
\texttt{model.\_moekv\_nd\_layer\_configs}.

\textbf{Hetero attention patch.} Per-layer cache position arrays and
step-count eviction trigger; see \texttt{moekv/heterogeneous\_attention.py}.

\end{document}